# Resting state fMRI functional connectivity-based classification using a convolutional neural network architecture

**Regina Meszlényi[1,2*], Krisztian Buza[2,3] and Zoltán Vidnyánszky[1,2]**

[1]Department of Cognitive Science, Budapest University of Technology and Economics, Budapest, Hungary

[2]Brain Imaging Centre, Research Centre for Natural Sciences, Hungarian Academy of Sciences, Budapest, Hungary

[3]Knowledge Discovery and Machine Learning, Rheinische Friedrich-Wilhelms-Universit¨at Bonn, Germany

**\* Correspondence:**
Regina Meszlényi
meszlenyi.regina@ttk.mta.hu

## Abstract

Machine learning techniques have become increasingly popular in the field of resting state fMRI (functional magnetic resonance imaging) network based classification. However, the application of convolutional networks has been proposed only very recently and has remained largely unexplored. In this paper we describe a convolutional neural network architecture for functional connectome classification called connectome-convolutional neural network (CCNN). Our results on simulated datasets and a publicly available dataset for amnestic mild cognitive impairment classification demonstrate that our CCNN model can efficiently distinguish between subject groups. We also show that the connectome-convolutional network is capable to combine information from diverse functional connectivity metrics and that models using a combination of different connectivity descriptors are able to outperform classifiers using only one metric. From this flexibility follows that our proposed CCNN model can be easily adapted to a wide range of connectome based classification or regression tasks, by varying which connectivity descriptor combinations are used to train the network.

## 1 Introduction

Resting state functional MRI (rs-fMRI) (Biswal et al. 1995) has become the most popular techniques for the investigation of the human brain's functional connectivity (B. Biswal et al. 2010; Greicius et al. 2003; Fox et al. 2009; Smith et al. 2011). Studying the resting state fMRI functional connectome offers a unique way to understand large scale functional organization of the human brain in health and disease. It has been shown recently that the efficacy of resting state fMRI network based classification can be improved substantially using machine learning techniques and raised the intriguing possibility of application of machine learning for fast and objective diagnosis of mental disorders such as autism (Abraham et al., 2017; Kassraian-Fard et al., 2016), schizophrenia (Arbabshirani et al., 2013; Kim et al., 2016), major depressive disorder (Rosa et al., 2015) and cognitive impairment (Liem et al., 2017).

Vast majority of these machine learning studies used traditional algorithms for classification, such as support vector machines (SVM) and least absolute shrinkage and selection operator (LASSO). In a recent review encompassing 77 MRI based machine learning papers (Brown and Hamarneh, 2016) more than half of the articles proposed to use SVMs. SVMs has many advantages especially in relatively small datasets, as these models can resist overfitting well (Cortes and Vapnik, 1995). Another popular method applies LASSO, which is a linear regression technique that enforces sparsity on its weights (Tibshirani, 1996). Sparsity has dual advantages, as it prevents overfitting in high-dimensional feature-spaces and creates well-interpretable results by selecting a relatively low number of important features. However recent developments in deep learning methods and theory shows that especially in case of complex high-dimensional datasets such as fMRI data, deep models have an exponential gain in efficiency over traditional machine learning models, i.e. from the same amount of training data, deep neural networks can learn exponentially more complex output function than traditional linear or kernel methods (Bengio et al., 2005, 2013; LeCun et al., 2015; Montúfar et al., 2014).

Arguably great potential lies in the application of deep learning techniques for fMRI based classification (Plis et al., 2014; Vieira et al., 2017). Indeed, fully connected deep neural networks were successfully applied for fMRI volume classification in (Jang et al., 2017) as well as for functional connectivity based classification (Kim et al., 2016). Furthermore, recent results show that convolutional neural network architectures might also be used for connectome based classification (Kawahara et al., 2017), which appears to be especially important in light of the remarkable success of convolutional neural networks in image classification and object recognition; e.g. at the famous ImageNet challenge (Krizhevsky et al., 2012; Szegedy et al., 2015).

However, successful application of deep convolutional networks for connectome based data classification holds a number of substantial challenges which must be overcome. An important requirement of deep learning techniques is the availability of a large amount of training examples (Krizhevsky et al., 2012). This appears to be a serious concern since in most structural and functional MRI studies, the number of measurements are very low (ranging from couple of dozens to a few thousands) as compared to the number of examples in datasets used for object recognition (e.g. the ImageNet's 1.2 million instances). Image classifiers may work extremely well with tens of neural layers and millions of trainable weights, however, small sample size results in overfitting in large networks, even with the most careful regularization. Therefore, in case of connectome based applications relatively simple convolutional network architectures should be designed, in correspondence with the amount of the available data.

Another key to the success of deep convolutional networks is the weight sharing (Bengio and Lecun, 1995) between different image regions. In particular, convolutional networks are able to learn properties of a given pixel's local neighborhood structure (usually 3x3 or 5x5 pixels), independent from the localization of that pixel. Sharing these weights simplifies the network and makes it robust against overfitting by reducing the number of trainable weights: in case of a fully connected layer design on an image with e.g. 128x128 pixels and 64 output neurons there are more than one million trainable weights, while a convolutional layer with 5x5 patches and 64 filters has only 1600. Additionally, the learnt representation becomes shift invariant, which is a crucially important factor in object recognition. With appropriate architecture or training design other types of invariance can be achieved like rotational or scale invariance (Marcos et al., 2016; Xu et al., 2014), that may also be beneficial for the results in several applications.

The task of classification based on brain connectivity data shows remarkable similarities to image classification. Both structural and functional connectomes can be represented as matrices, where each row and each column corresponds to a voxel, or in most cases a brain region (ROI) from a given parcellation scheme, and the value of the (i,j)-th entry of this matrix describes the connectivity between the i-th and j-th brain region. These matrices can be treated as images, with matrix entries analogous to pixels. However, the structure of the local neighborhood in connectome data is not equivalent to traditional image datasets, patterns we try to recognize in this case are by no means shift invariant, and local neighborhoods (3x3 or 5x5 pixel patches) mean little as the ordering of ROIs is not necessarily interpretable. In connectome data based learning one should consider the graph structure behind the connectivity matrix to determine how to share weights, i.e. as brain graphs are usually fully connected, the neighborhood of one ROI contains every other region, and convolutional filters should be designed to take this into account.

Application of convolutional architectures to connectome data is in its very early stage (for review see Vieira, Pinaya, and Mechelli 2017). In a recent study (Kawahara et al., 2017), the authors proposed three types of convolutional filters for structural connectome data, edge-to-edge, edge-to-node and node-to-graph filters, which use different combinations of row and column wise convolutional filters. Their BrinNetCNN architecture takes tractography-based (structural) connectivity matrices as input, and is used to estimate continuous variables like subject age and behavioral metrics. However, an important feature of convolutional network architectures was not exploited in these study. Convolutional networks were designed to be able to treat not only single grayscale images, but to combine information from color (RGB) channels (Krizhevsky et al., 2012; LeCun et al., 2015): the different channels hold information about different aspects of the same "pixel", and this information is smartly combined together with learned weights in the network. Similarly these networks can straightforwardly combine connectivity matrices of different metrics, while keeping the information about which features correspond to the same ROI pairs (i.e. which pixel) by treating connectivity matrices as color channels in standard image processing tasks.

In the present study we aimed at investigating the application of convolutional networks for functional connectome classification using a simple connectivity fingerprint-based convolutional filter. In our CCNN model we treated one ROIs whole connectivity fingerprint, i.e. one row (or column) of the matrix, as a unit, so that those weights can be shared across the whole connectivity matrix. The rationale behind our approach is the following: if we assume that some (or many) regions show altered connectivity between the two classes we try to differentiate, the learned convolutional filter will distribute large weights to those ROIs, i.e. when we convolve every ROIs' fingerprint (every row of the matrix) with the filter, the connectivity strength with those altered regions will have large influence on the output. Our proposed filters output one value for each ROI's connectivity fingerprint and the number of trainable weights equals to the number of ROIs. Based on (Kawahara et al., 2017) one output value per ROI can be calculated by sequentially applying an edge-to-node and a node-to-graph filter to the connectivity matrix. However using this filter combination requires the training of three times as many weights as ours, largely increasing the size the neural network model that may lead to the aforementioned disadvantages (overfitting) when the number of training examples is low.

In addition to implementing our convolutional filter, we also tested the hypothesis that combination of different functional connectivity metrics will improve classification accuracy. We assumed that even though using combined inputs naturally increase the number of trainable weights of the model, adding new sources of information might still increase classification performance. Traditionally, functional connectivity strength is measured using correlation coefficient calculation. However, several additional methods have been proposed, that can grasp the dynamic properties of functional connectivity (Chang and Glover, 2010; Fransson and Marrelec, 2008; Salvador et al., 2005). For the purpose of classification it is desirable that the connectivity matrix is stable between measurements, but it has been shown that correlation coefficients between resting state time-series are dynamically changing due to alternating brain states (Allen et al., 2014; Chang and Glover, 2010; Chen et al., 2015). To overcome this problem, we have recently proposed Dynamic Time Warping (DTW) distance (Meszlényi et al., 2016b, 2017) and warping path length (Meszlényi et al., 2016a) as new metrics for functional connectivity strength and phase stability. As DTW distance can handle and correct for dynamically changing phase-relationships between brain regions, we were able to demonstrate that DTW distance based connectivity strength calculation indeed results in stable connectivity between measurements (Meszlényi et al., 2017), while the length of the warping path provides important additional information about connection stability (Meszlényi et al., 2016a). In our previous papers we also showed that both DTW distance and warping path length can be successfully used for connectivity based classification (Meszlényi et al., 2016a, 2016b, 2017). As DTW distance and warping path length describe two complementary aspects of connectivity, in the current study we examine the effect of combination of different functional connectivity metrics on the classification accuracy of the convolutional network by combining input from these two measures, and we compare its classification performance to results based on single connectivity features calculated with three metrics: correlation, DTW distance and warping path length.

In this paper, we demonstrate the feasibility of our convolutional model, the CCNN on a simulated dataset and test our proposed approach on a publicly available datasets for amnestic mild cognitive impairment (aMCI) classification and compared its performance to a traditional neural network with one hidden layer and a deep neural network. We demonstrate that the connectome-convolutional network architecture outperforms both the simple neural network and the deep model in all cases, while the best performing models are those CCNNs that use a combination of different connectivity metrics.

## 2 Materials and methods

### 2.1 Amnestic mild cognitive impairment dataset

We used publicly available data from Consortium for Reliability and Reproducibility (CoRR) (Zuo et al., 2014): the LMU 2 and 3 dataset (Blautzik et al., 2013a, 2013b). The datasets contain forty-nine subjects who are older than 50 years (22 males, age (mean ± SD): 68.6 ± 7.3 years, 25 diagnosed with aMCI), each subject participated at least two 366-sec-long resting-state fMRI measurements, the subjects have 146 resting-state measurements in total.

The dataset was collected at the Institute of Clinical Radiology, Ludwig-Maximilians-University, Munich, Germany, on a 3 T Philips Achieva scanner (Best, The Netherlands). High-resolution anatomical images were acquired for each subject using a T1-weighted 3D TFE sequence (1 mm isotropic voxels; TR = 2400 ms; FOV = 256 mm; acceleration factor = 2). A total of 120 functional images over 366 sec were collected with a BOLD-sensitive T2* weighted GRE-EPI sequence (4 mm slice thickness with 3 mm x 3 mm in-plane resolution; TR = 3000 ms; TE = 30 ms; FOV = 192 mm). 28 axial slices were acquired in ascending acquisition order covering the whole brain. Further details are available on the website of the datasets (http://fcon_1000.projects.nitrc.org/indi/CoRR/html/ lmu_2.html and http://fcon_1000.projects.nitrc.org/indi/CoRR/html/ lmu_3.html).

### 2.2 Preprocessing

Preprocessing of the imaging data was performed using the SPM12 toolbox (Wellcome Trust Centre for Neuroimaging) and custom-made scripts running on MATLAB 2015a (The MathWorks Inc., Natick, MA, USA). Each subject's functional images were motion-corrected, the T2* images from all sessions were spatially realigned to the mean T2* image. Then, EPI images were spatially smoothed using a 5 mm full-width half maximum Gaussian filter. The anatomical T1 images were coregistered to the mean functional T2* images used in the realignment step. The coregistered T1 images were segmented using the unified segmentation and normalization tool of SPM12. The resulting grey matter (GM) mask was later used to restrict the analysis of the T2* images to GM voxels; while the white matter (WM) and cerebrospinal fluid (CSF) masks were used to extract nuisance signals that are unlikely to reflect neural activity in resting-state time-series. The realigned and coregistered images were normalized to the MNI-152 space using the transformation matrices generated during the segmentation and normalization of the anatomical images. After regressing out the head-motion parameters, the mean WM and the mean CSF signals, residual time courses from all GM voxels were band-pass filtered using a combination of temporal high-pass (based on the regression of ninth-order discrete cosine transform basis set) and low-pass (bidirectional 12th-order Butterworth IIR) filters to retain signals only within the range of 0.009 and 0.08 Hz (Meszlényi et al., 2017).

## 2.3 Functional connectivity calculation

To calculate ROI-based whole-brain functional connectivity we used the Willard functional atlas of FIND Lab, consisting of 499 functional regions of interest (Richiardi et al., 2015) to obtain 499 functionally meaningful averaged BOLD signals in each measurement. From this 499 time series we can calculate full connectivity matrices leading to 499x498/2 = 124251 independent pairwise connectivity features.

Functional connectivity can be characterized with various metrics including traditional correlation coefficient, Dynamic Time Warping distance (Meszlényi et al., 2016b, 2017), and warping path length (Meszlényi et al., 2016a). For a brief description of the DTW based methods see Supplementary material.

## 2.4 Simulated dataset

To demonstrate the strengths of our proposed convolutional filters, we created an artificial dataset of connectome matrices. As a base connectome we choose a random correlation based functional connectome of a healthy subject and we created three modified versions of this healthy connectome based on a connectome matrix of a patient with aMCI. We generated modifications by replacing the rows and columns corresponding to randomly chosen ROIs of the healthy connectome with the rows and columns of the aMCI connectome, specifically we created connectomes with one, five and ten ROIs replaced.

From the unchanged healthy connectome and a modified connectome we created 75-75 replicas and added random Gaussian noise to the connectomes to generate 150 unique instances, taking into account that the matrices have to stay symmetrical (i.e. we randomly generated noise matrices and symmetrized them by adding its transpose to it). We added noise with different weights, i.e. we normalized the noise values to have a maximal absolute value of one (standard deviation equals to 0.17) and we added noise with weights ranging from one to ten. With the three modification levels and ten weight-levels of added noise we created altogether thirty simulated datasets.

## 2.5 Classification

We aimed to classify simulated datasets and amnestic mild cognitive impairment based on functional connectivity data. To estimate classification performance, we applied cross-validation. In the aMCI dataset, there are 146 instances, but measurements of the same subjects are not independent, therefore we took this into account during cross-validation. In this dataset, we have measurements from 49 subjects, so we applied a 7-fold cross-validation: we randomly divided the 49 subjects to seven folds, and each fold contains all the measurements of the subjects assigned to the fold. We used this same partitioning to evaluate all classifiers. In the simulated datasets we have 150 unique instances therefore we applied a simple 10-fold cross-validation.

We asses classification performance primarily with accuracy (i.e., proportion of correctly classified instances) as the classes are balanced in both of our classification tasks, but to present more detailed information, we calculated the area under the receiver operator characteristics curve (AUC) as well.

### 2.5.1 Connectome-convolutional neural network

To achieve better classification performance we designed a novel convolutional network architecture, the CCNN for functional connectivity pattern classification. Traditional convolutional networks (Krizhevsky et al., 2012; LeCun et al., 2015) usually apply square weight patches (3x3 or 5x5 filters) for convolution, as for image classification, important information is contained in square neighborhoods of pixels. For functional connectivity classification, we arranged the connectivity features into 499x499 matrices (as we consider 499 ROIs), and we apply convolution in two layers, first line-by-line (1x499 convolution filters), than by column (499x1 filters). The second convolution layer provides input for a fully connected hidden layer with 96 neurons that feeds two output neurons corresponding to the two classes (see Fig. 1. C).

In the first convolutional layer we train 64 filters, while in the second convolutional layer has 128 filters. This means that the first convolutional layer extracts 64 features per ROI, i.e. we calculate 64 differently weighted sums of each ROI's connectivity fingerprint. The second layer reduces the dimensionality further: it outputs 128 feature for each instance, and this 128 dimensional feature vector serves as input for the fully connected layer. As we have around 150 instances in both datasets, this means that the number of extracted features is approximately matched to the number of instances. In the convolutional neural network, we applied rectified linear unit (ReLU) (Nair and Hinton, 2010) non-linearity and on the output layer we apply the softmax function (Bridle, 1990) to calculate the probability of each instance belonging to a class. In binary classification, we could use only one output neuron to determine the probability of an instance belonging to the positive class, however our adopted approach with one neuron per class and a softmax function can be straightforwardly implemented for multiclass classification tasks as well. To train a robust classifier, we applied drop-out regularization (Srivastava et al., 2014; Wager et al., 2013) with keeping probability of 0.6 and an Adam optimizer (Kingma and Ba, 2014).

In case of combined CCNN classifiers the input consists of two 499x499 matrices of connectivity features, each of which can be considered as a "channel", and we apply the convolution to both "channels", i.e. to the two matrices simultaneously, similarly to how convolutional layers work on the RGB channels of colored images. With this approach we can explicitly inform the network which connectivity features belong to the same ROI pair, so the algorithm can take advantage of this additional information as well. It is worth to note that the size of the CCNN increases less than one percent with the addition of a new channel. The original (one-channeled) network has altogether 499*64+499*64*128+128*96+96*2 = 4,132,224 trainable weights plus 290 biases, while with two channels, the CCNN has 2*499*64+499*64*128+128*96+96*2 = 4,164,160 trainable weights plus the same 290 biases.

The proposed connectome-convolutional neural network was implemented in Python using TensorFlow, and the source code of the model is available on GitHub (https://github.com/MRegina/connectome_conv_net).

### 2.5.2 Baseline classification

As a simple baseline of random classification, we applied a binomial method described in (Pereira et al., 2009). In two class classification, the random classifier has p=50% chance of predicting the true label, therefore the probability of obtaining not more than k correct labels out of n trials can be calculated from the cumulative binomial distribution function (Eq. 1):

$$\text{Eq. 1.:} \quad F_{Binom}(n, k, p) = \sum_{i=0}^{k} \binom{n}{i} p^i \cdot (1-p)^{n-i}$$

For threshold of significance, we choose the 95 percentile, i.e. we searched for the k value, where $F_{Binom}(n, k, 0.5) \approx 0.95$, from that the baseline accuracy can be calculated as $k/n$. In case of the simulated dataset the calculated baseline accuracy is 56.67% with $F_{Binom}(150,85,0.5) = 0.959$, while for the aMCI dataset the threshold of significance is 56.85% with $F_{Binom}(146,83,0.5) = 0.959$.

### 2.5.3 Simple neural network classifier

For a stricter baseline classification result on the available connectivity datasets, we created a traditional neural network (Widrow and Lehr, 1990), containing an input layer of 124251 input neurons (the number of independent functional connectivity features), one hidden layer with 128 neurons (equal to the number of features extracted by the CCNN) and two output neurons for the two classes (See Fig. 1. A). The hidden layer has a sigmoid nonlinearity and on the output layer we apply a softmax function to calculate the probability of each instance belonging to a class. The network is trained with stochastic gradient descent and cross-entropy as loss function (Goodfellow et al., 2016). From now on we will refer to this network as the simple neural network.

The number of trainable weights in this network equals to 124251*128+128*2 = 15,904,384 plus 130 biases, almost four times more than in the CCNN model. In case of combined classifiers, where we aim to learn from two connectivity descriptors, thus the number of input neurons equals to 2*124251=248502, therefore the number of trainable weights nearly doubles: 248502*128+128*2 = 31,808,512 plus the 130 biases.

### 2.5.4 Deep neural network classifier

To demonstrate how our convolutional architecture performs compared to a state-of-the-art neural network architecture, we created a multi-layered (deep) neural network. The input layer is similar to the simple neural network, it consists of 124251 or 2*124251 neurons depending on whether we use data from a single connectivity descriptor, or we combine two metrics. The first hidden layer has 128 neurons, i.e. this layer extracts 128 features per instance, similar to the convolutional layers in our CCNN architecture. The second hidden layer contains 96 neurons similarly to the convolutional networks' fully connected layer, and lastly the multi-layered neural network has the same two output neurons as both the simple and the convolutional architecture (see Fig. 1. B). In this neural network classifier we applied ReLU nonlinearities in the hidden layers and softmax at the output layer, and drop-out regularization with 0.6 keeping probability and the Adam optimizer similarly to the convolutional neural network, i.e. the primary source of the differences between these two models is the convolutional architecture. As the neural network is multi-layered and we applied all the aforementioned deep learning techniques in its training, we will refer to it as the deep neural network classifier throughout the paper.

The number of trainable weights in the deep neural network is 124251*128+128*96+96*2 = 15,916,608 plus 226 biases, slightly more than in the simple neural network. In case of combined input data the number of trainable weights almost doubles here as well: 248502*128+128*96+96*2 = 31,820,736 plus the 226 biases.

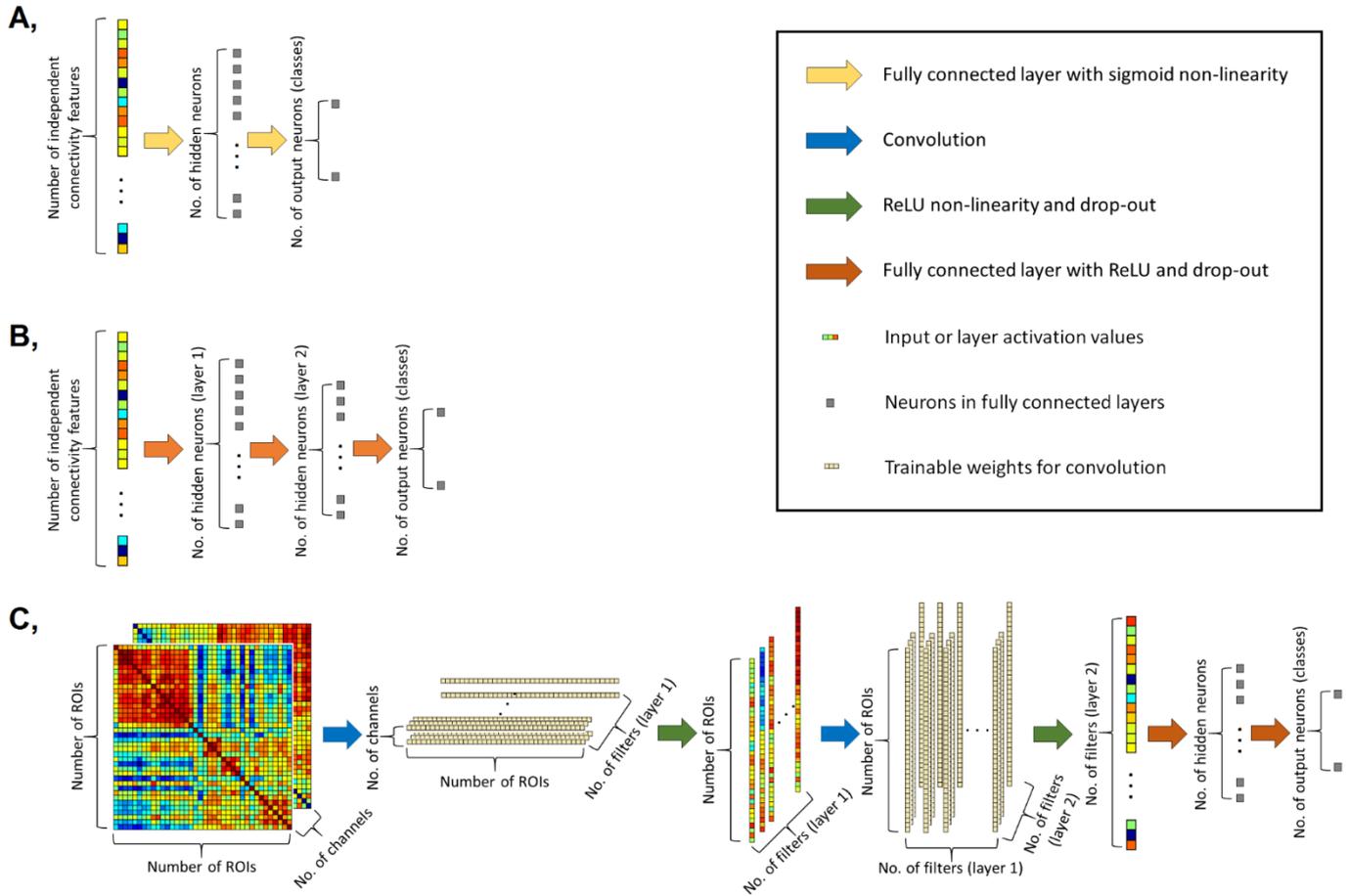

*Figure 1.: A, Architecture of the simple neural network classifier B, Architecture of the deep neural network classifier C, Architecture of the proposed connectome-convolutional neural network model*

## 3   Results

In the results section we describe classification results with two performance metrics: accuracy and area under the receiver operator characteristics curve (AUC). To determine if the difference between two classifiers' performance is significant, we applied a binomial test (Salzberg, 1997), we consider the difference significant if the calculated p-value is lower than 0.05. We also visualized what the best performing classifiers learned from the available data.

### 3.1   Classification of simulated data

On Fig. 2 we summarized the classification accuracies achieved by a simple neural network, the deep network and our proposed connectome-convolutional architecture as a function of the number of modified ROIs and the level of noise. As expected, the results revealed that classification performance of the simple neural network increases with the number of modified ROIs, while it decreases as the weight of added noise increases. The fully connected deep neural network can clearly outperform the simple architecture in most cases, but interestingly it shows below or near random performance when the noise level is relatively low, likely due to overfitting. The results clearly show that the CCNN architecture has the best overall performance, significantly outperforming even the deep neural network in several cases. As weight sharing considerably reduces the number of trainable weights, the connectome-convolutional neural network does not suffer from overfitting at low noise levels: the CCNN achieves 100% accuracy at low noise levels.

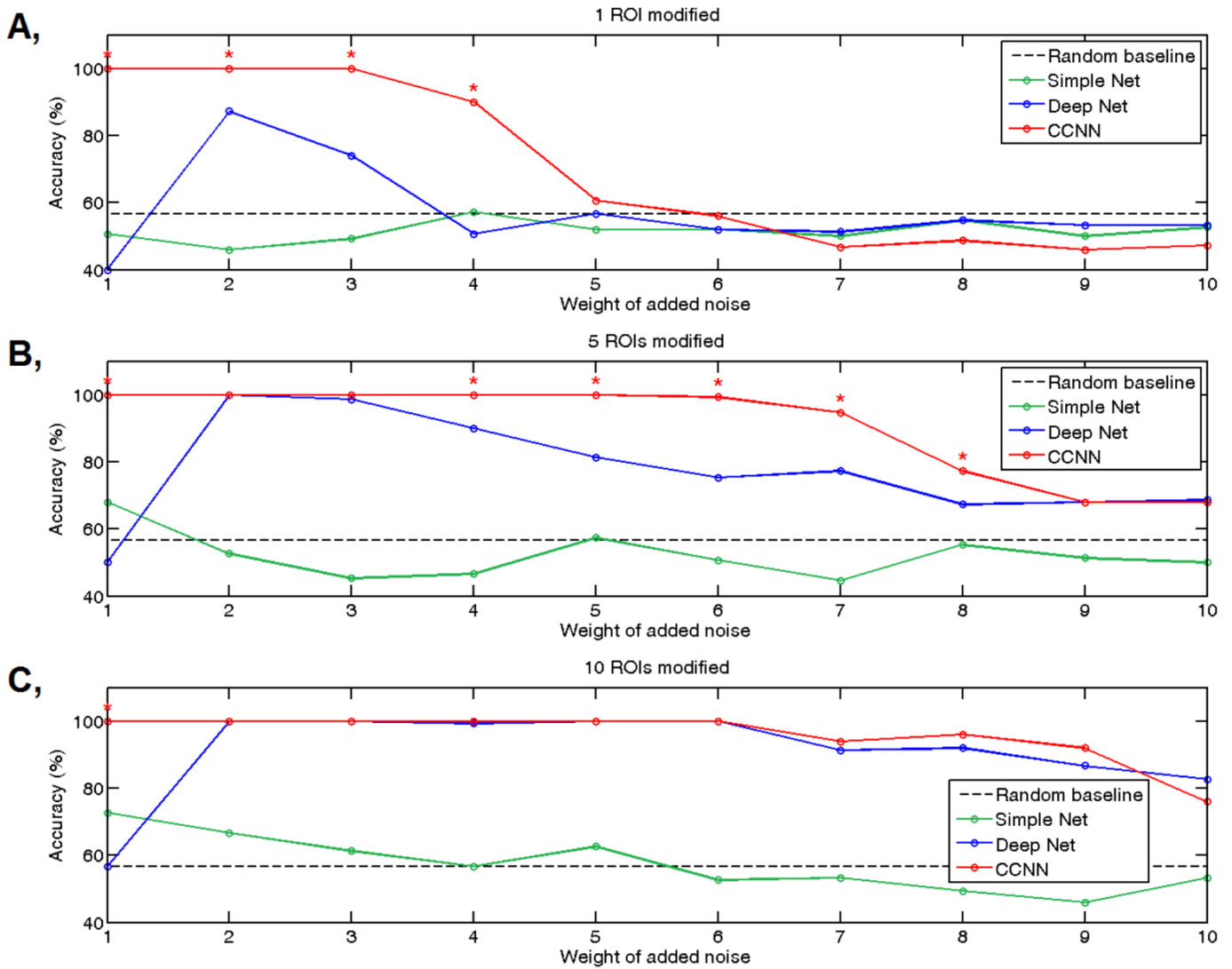

*Figure 2.: Accuracies of classification of simulated data with the simple (green), deep (blue) and connectome-convolutional (red) neural networks. The black dashed line represents the random baseline, significant (p<0.05) differences between the deep and the connectome-convolutional networks' results are denoted with red stars. A, Classification accuracies on the dataset with one ROI modified B, Classification accuracies on the dataset with five ROIs modified C, Classification accuracies with ten ROIs modified*

Besides demonstrating the connectome-convolutional neural network's remarkable performance and robustness, we also showed that based on the first layer's weights in these neural networks we can indeed recover which ROIs played important role in the classification. We investigated our hypothesis that the learnt convolutional filters should distribute high absolute value weights to the ROIs which behave differently between classes, i.e. those ROIs that have the largest sum of absolute values through the first convolutional layer's 64 filters should overlap with the ones that were actually modified in the simulated datasets. Naturally we should keep in mind that as the CCNN has more than one layers, the fact that some ROIs may not have a large sum of weights in the first layer does not mean that they do not play significant role in the classification, as filter outputs are further weighted in the next layer. We evaluated our hypothesis at the noise level of five, where still all CCNNs are able to classify the data better than random, but the added noise has a large weight. Our experiment showed that based on the dataset, where only one ROI was modified between the classes, the connectome-convolutional architecture distributed by far the largest absolute values to this altered ROI. In case of the dataset where five ROIs were modified, the CCNN model identified four of these among the five ROIs with the largest sum of absolute weight. In the dataset with ten ROIs altered, also four ROIs could be recovered. The learnt weights of the first layer of the connectome-convolutional networks are visualized in Supplementary Fig. 2.

### 3.2 Amnestic mild cognitive impairment classification

Table 1. shows the performance of the examined neural networks for amnestic mild cognitive impairment classification - the simple neural networks, deep neural networks and connectome-convolutional neural networks - using various feature sets: pairwise correlation coefficients, DTW distances and warping path length features, and the combination (i.e., union) of the latter two feature sets.

| Simple Net | CORR | DTW | Path length | DTW+Path length |
|---|---|---|---|---|
| Accuracy (%) | 50 | 52,1 | 57,3 | 56,2 |
| AUC | 0,515 | 0,505 | 0,59 | 0,588 |
| Deep Net | CORR | DTW | Path length | DTW+Path length |
| Accuracy (%) | 50,7 | 61,6 | 62,3 | 61,0 |
| AUC | 0,533 | 0,634 | 0,635 | 0,611 |
| CCNN | CORR | DTW | Path length | DTW+Path length |
| Accuracy (%) | 53,4 | 65,1 | 64,4 | 71,9 |
| AUC | 0,521 | 0,684 | 0,672 | 0,746 |

*Table 1.: Performance measures of the examined neural networks based on correlation, DTW distance, DTW path length, and the combination (i.e., union) of the latter two feature sets.*

First we compared classification performances to the random baseline, i.e. the threshold of significance in accuracy is 56.85%. Only Dynamic Time Warping based measures did outperform this threshold, i.e. path length based classification achieved significant results with the simple, deep and our CCNN architecture, and DTW distance based classification was successful with the deep and the connectome-convolutional neural network. It is also interesting to compare whether the differences in the performance of the simple neural model and the CCNN are significant. We found no difference in case of the correlation based classifiers (p=0.32), and also in case of the path length based classifier (p=0.15) although in this case the CCNN's accuracy is substantially (7%) higher than that of the simple neuronal network. The difference in classification performance was significant in case of the DTW distance based classifiers with p=0.012. When comparing results of the deep and the connectome-convolutional neural network, we found that even though the CCNN systematically outperforms the deep model, these differences are not significant (p>0.2 in all three cases). Out of the classifiers based on single connectivity feature sets, the best classification performance was achieved by DTW distance-based CCNN model.

Next we tested whether training connectome-convolutional neural networks using the combination of the connectivity feature sets based on DTW distance and warping path length leads to better classification performance compared with the previous models. The combined CCNN model achieved higher classification performance than the threshold of random classification. It significantly outperformed the simple neural network trained on combined features (p=0.0034) and most importantly the difference between the deep and the connectome-convolutional neural networks' results also showed high significance (p=7.7e-4). We also compared the performance of the CCNN trained on combined data with the performance of CCNNs trained on only one type of connectivity features. Classification performance of the combined model was significantly higher than that of the warping path length based connectome-convolutional network (p=0.031), but did not differ significantly from the DTW distance based model's performance (p=0.092). With the results of the combined models, the overall best classification performance was achieved by the CCNN trained on both DTW distance and warping path length data.

As we demonstrated with the simulated dataset, from the weights of the connectome-convolutional neural network, we can identify which ROIs played important role in the classification. As the combined DTW distance and path length based CCNN model achieved the overall best performance, we analyzed the weight distribution of this model, i.e. the 2*64*499 weights of the first convolutional layer, from which the first 64*499 weights correspond to DTW distance features, while the second 64*499 weights correspond to warping path length values. For the sake of biological interpretation we aimed to incorporate all meaningful information to the learned weights, i.e. we trained our CCNN architecture on the combined feature set of the whole dataset.

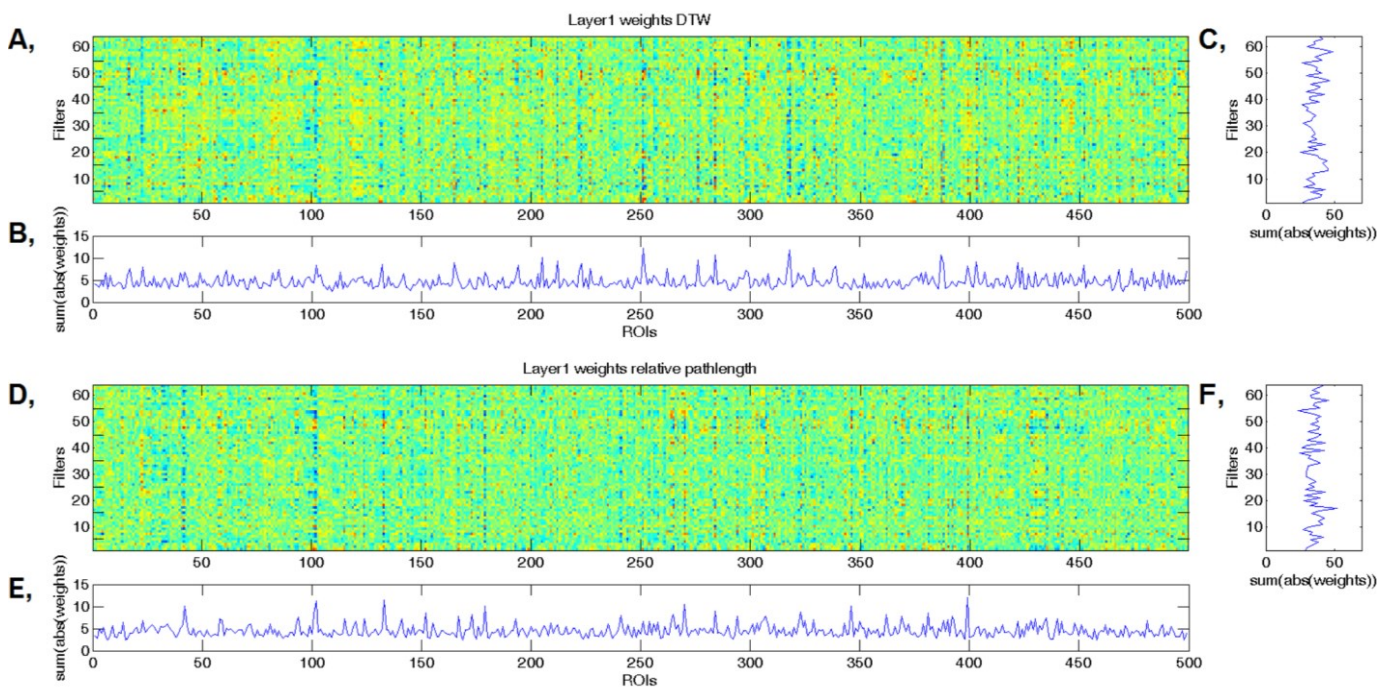

*Figure 3.: Learned weights of the first convolutional layer of the combined CCNN model trained on the whole aMCI dataset. This layer has 2*64*499 weights, and we present the first 64*499 weights corresponding to DTW distance values (A-C) separately from the second 64*499 weights that correspond to warping path length features (D-F). A, Colormap of the first 64x499 weights of the first convolutional layer of the combined CCNN model corresponding to DTW distance features. B, To determine which ROIs play important role in the classification, we summarized the absolute values of the weights through the 64 filters. High values represent ROIs that have significant effect in most filters. C, To determine which filters are the most effective, we summarized the absolute values of weights through the 499 ROIs. High values represent filters that have substantial influence on the output. D,*

*Colormap of the second 64x499 weights of the first convolutional layer of the combined CCNN model corresponding to relative warping path length features. E, Summarized absolute values of the weights through the 64 filters. High values represent ROIs that have significant effect in most filters. F, Summarized absolute values of weights through the 499 ROIs. High values represent filters that have substantial influence on the output.*

In Fig. 3. B and E it is clearly visible that some ROIs have high absolute value weights in most filters, i.e. connectivity with these brain areas has an important effect on the output of the first layer. We note that the most important ROIs of DTW and warping path length have no significant overlap. Although DTW distance and warping path length can correlate, they measure substantially different aspects of co-activity: DTW distance measures connectivity strength, whereas waring path length quantifies the stability of the connection, i.e., how frequently the phase difference changes between ROIs. From the DTW distance based features the right amygdala and hippocampus, bilateral caudate and left putamen, anterior and medial cingulate cortex, dorsolateral prefrontal cortex, temporal and occipital regions show alterations between the two classes (Fig. 4. A). From path length features, the medial regions of bilateral thalamus, left caudate, posterior cingulate cortex and precuneus, dorsolateral prefrontal cortex, insula and other occipital and frontal areas play important role in classification according to aMCI (Fig. 4. B).

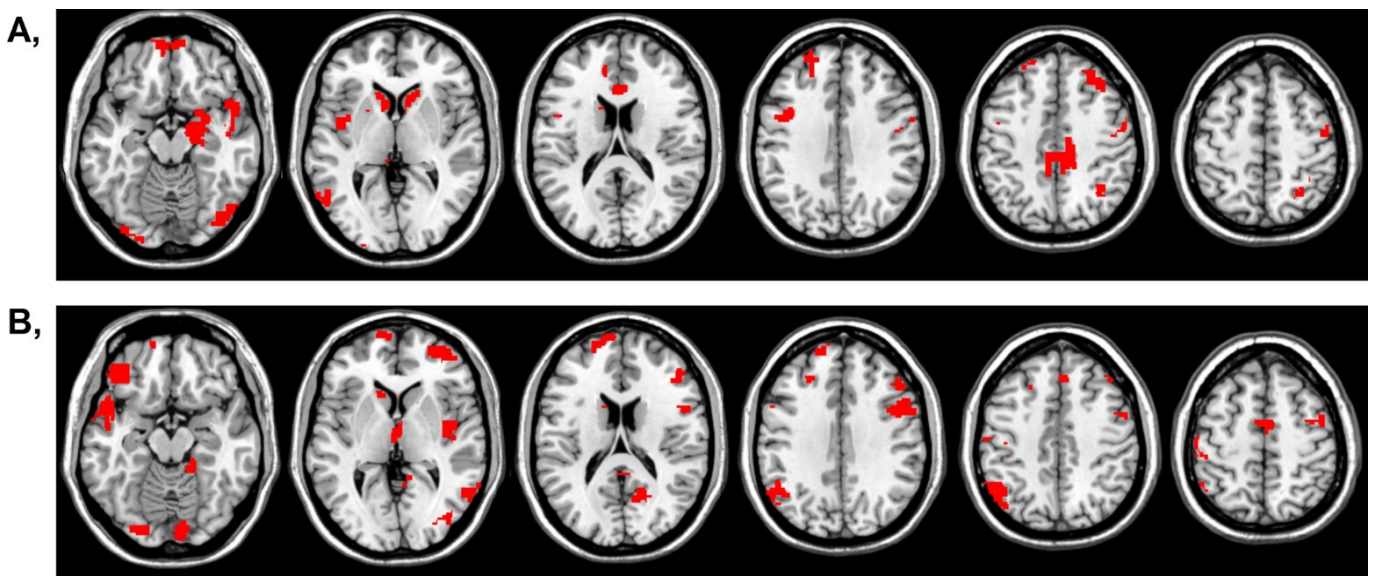

*Figure 4.: Most influential ROIs based on the first convolutional layer's weights for aMCI classification with CCNN. A, Important ROIs based on DTW distance features B, Important ROIs based on warping path length features*

## 4 Discussion

The results of the present study clearly show that using convolutional neural architectures for connectome classification has great potential, even in case of relatively small sample sizes. We demonstrated that our proposed CCNN architecture can significantly outperform not only a traditional neural network, but a deep neural network model as well. With the simulation study we were able to prove that the connectome-convolutional network is much less prone to overfitting than the deep model while it systematically outperforms both deep and simple neural architectures at different noise and modification levels. We also showed that by analyzing which brain regions got the largest absolute weights in the convolutional filters of the first layer, we can indeed recover ROIs that contained information relevant for the classification (i.e. those ROIs that were truly modified in the simulated dataset).

In amnestic mild cognitive impairment classification the CCNN model also systematically outperformed the deep and simple neural networks and most importantly, the connectome-convolutional network was able to utilize information from multiple different connectivity metrics. In this case the difference between the CCNN model and the second best performing deep neural network was highly significant, which is most likely the result of the fact, that by doubling the number of input features, the size of the deep neural network also doubles, while the size of the CCNN architecture only slightly increases (less than one percent in our particular case). Consequently convolutional networks are less prone to overfitting and can exploit additional information more efficiently (Goodfellow et al., 2016). This feature has great potential in future research, as the CCNN model may not only benefit from different functional connectivity metrics, but it can for example straightforwardly combine structural and functional connectivity information as well.

For a thorough comparison we also performed experiments with traditionally well performing machine learning models. Due to the extremely high number of features in the combined DTW distance and warping path length dataset, we applied two methods that accomplish feature selection: linear SVM classifier combined with ANOVA F-test based feature selection (Abraham et al., 2014) and a LASSO model (Meszlényi et al., 2016a). The results of these experiments (see Supplementary Material) showed that both SVM and LASSO can perform better than the deep neural network given the relatively low number of samples, however neither methods achieved better results than our proposed CCNN architecture. Although the best performing approach (a LASSO classification) achieved 2% less accuracy than the CCNN model, we argue that our proposed method holds much greater promises than this small performance gain. As the sample size in fMRI experiments continues to increase (Smith et al., 2014) and multimodal paradigms also gain popularity (Brown et al., 2012), the gap between the performances of the CCNN model or similar deep learning based techniques and traditional machine learning methods bound to increase (LeCun et al., 2015).

Amnestic mild cognitive impairment classification performance based on DTW distance that integrates dynamic connectivity and DTW warping path length that describes phase-stability was significantly higher than that based on the correlation coefficients. This is in agreement with recent findings showing strong alterations in dynamic connectivity and connection stability in Alzheimer's disease and mild cognitive impairment (Córdova-Palomera et al., 2017). The best performance was achieved by a CCNN that was trained on a combination of these two metrics, namely DTW distance and path length. The most influential regions of the classification include regions of the default mode network (e.g. posterior cingulate cortex and precuneus) and executive control network (e.g. dorsolateral prefrontal cortex) as well as several subcortical regions, including the hippocampus, basal ganglia and amygdala. These results are in close agreement with previous research providing converging evidence for the abnormalities in resting-state functional connectivity of these regions in amnestic mild cognitive impairment (Lau et al., 2016; Liang et al., 2011, 2015; Yang et al., 2012).

Our previous findings have shown that classification based on DTW distance and warping path length can outperform a correlation based paradigm in different datasets and even with different classifiers and classification targets (Meszlényi et al., 2016a, 2016b, 2017). Our results presented in this paper also confirm that taking into account the dynamic properties of functional connectivity can assist classification. DTW distance and path length based classifiers did outperform the correlation based models, while the best results were achieved by combining connectivity features of these two metrics. To understand why the combination of these connectivity descriptors leads to better results, it is important to notice how DTW distance and warping path length are related. In case of very strong connections, i.e. when the two compared time-series are very similar almost no warping (time-series elongation) is necessary thus the length of the warping path will be short, while the calculated DTW distance will be very close to zero. Additionally in case of independent random time-series the DTW distance is large as the two series cannot be meaningfully matched, also independent time-series mean more editing steps, therefore the length of the warping path increases as well (Tomasev et al., 2015). However, there are connections, where the Dynamic Time Warping algorithm can generate a good match between the two time-series, but the phase relationship between them is unstable, i.e. the time-delay structure is dynamically changing through the measurement. This type of relationship can result in long warping path lengths despite the relatively low DTW distance values, and these are the connections where warping path length contains interesting additional information (Meszlényi et al., 2016a).

Naturally deep learning techniques and particularly our CCNN method have their drawbacks as well. Deeper networks take longer time to train than traditional shallow neural networks or other methods like SVMs, however with modern deep learning frameworks and GPU computing our CCNN model can be trained in sevenfold cross-validation within an hour. Another difficulty is the selection of hyper-parameters. Deep networks have several architectural parameters like the number of different convolutional and fully connected layers, the number of filters and neurons in each layer or the activation function, as well as training parameters like initialization, loss-function, learning-rate or optimization function. Due to the long training time of these models, thorough hyper-parameter learning is usually not feasible, typically only a small number of parameters can be tuned, while most parameters have to be set based on experience (Goodfellow et al., 2016; LeCun et al., 2015). We also note that even though the convolutional architectural design can significantly decrease the number of trainable parameters compared to fully connected deep networks, the number of these trainable weights can still be very high compared to the number of samples in the dataset. Therefore careful regularization (e.g. with dropout) is essential to the success of these models, and the effectiveness of their application to datasets with extremely few training examples (e.g. less than a hundred measurements) is debatable.

In this paper we presented a connectome-convolutional neural network architecture that was designed to be able to analyze bran connectivity matrices and classify subject groups based on the connectivity fingerprints of brain regions. With an experiment on simulated datasets we showed that besides having high classification performance, the CCNN architecture we implemented can identify ROIs that have altered connectivity strength values. On a real-world dataset of healthy elderly controls and patients with amnestic mild cognitive impairment we were also able to demonstrate that the connectome-convolutional neural network can effectively utilize information from multiple functional connectivity descriptors. Namely the overall best classification accuracy was achieved by the CCNN model trained on a combination of Dynamic Time Warping distance and warping path length connectivity matrices. The brain regions that had large influence on the classification results are well aligned with the current research findings on amnestic mild cognitive impairment. From these results we can conclude that the presented connectome-convolutional neural network architecture should be considered as an efficient tool for brain connectivity-based classification tasks, especially in experiments where multiple connectivity descriptors, i.e. different functional connectivity measures or functional and structural connectivity information is available.

## 5 Conflict of Interest

*The authors declare that the research was conducted in the absence of any commercial or financial relationships that could be construed as a potential conflict of interest*.

## 6 Author Contributions

All coauthors contributed to this study: RM, KB, ZV

## 7 Funding

This work was supported by a grant from the Hungarian Brain Research Program (KTIA_13_NAP–A–I/18) to Zoltán Vidnyánszky. Krisztian Buza was supported by the National Research, Development and Innovation Office - NKFIH PD 111710 and the János Bolyai Research Scholarship of the Hungarian Academy of Sciences.

# 9 Supplementary Material - Dynamic Time Warping distance and warping path length

Dynamic Time Warping is a time-series distance measure that can correct for even dynamically changing phase-differences between signals. DTW was first applied in the field of speech recognition (Sakoe and Chiba, 1978). DTW distance can be efficiently used for time-series classification (Ding et al., 2008; Xi et al., 2006), and recently our group has demonstrated its potential for fMRI data analysis (Meszlényi et al., 2016a, 2016b, 2017).

DTW is a so called edit distance, which means that it measures the "cost" of transforming one time-series to the other one (Suppl. Fig. 1. A). Two editing steps are possible for transforming two time-series $x_1$ and $x_2$: replacement of an element of $x_1$ to an element of $x_2$ or an elongation of an element in $x_1$ or $x_2$. The cost of each editing step is the difference of the matched elements, while the overall cost of transformation is the sum of the costs of each editing step: DTW distance is the minimal possible transformation cost. We can calculate DTW distance of $x_1$ and $x_2$ with length of $l_1$ and $l_2$ by filling the entries of an $l_1$ x $l_2$ matrix column-by-column or row-by-row (Suppl. Fig. 1. B,C) based on Equation 1:

Eq 1.: $$\mathrm{DTW}(i,j) = \begin{cases} \|x_1(i), x_2(j)\| + \min\{\mathrm{DTW}(i, j-1), \mathrm{DTW}(i-1, j), \mathrm{DTW}(i-1, j-1)\} & \text{if } i,j > 1 \\ \|x_1(i), x_2(j)\| + \mathrm{DTW}(i, j-1) & \text{if } i = 1, j > 1 \\ \|x_1(i), x_2(j)\| + \mathrm{DTW}(i-1, j) & \text{if } j = 1, i > 1 \\ \|x_1(i), x_2(j)\| & \text{if } i = 1, j = 1 \end{cases}$$

where $x_1(i)$ denotes the *i*-th value in time series $x_1$ and $x_2(j)$ denotes the *j*-th value in time series $x_2$ and $DTW(l_1,l_2)$ is the DTW-distance of the two time series (Suppl. Fig. 1. D). To constrain the maximal allowed phase-difference (time-shift between matched elements) and to speed-up DTW-calculations, we can calculate only those entries of the DTW matrix that are close to the main diagonal (Sakoe and Chiba, 1978). The maximal allowed time-shift is the so called warping window (Suppl. Fig. 1. B).

After filling-in the DTW matrix, one can reconstruct which editing steps led to the minimal DTW distance value, i.e. we can determine the optimal matching between each element of the two time series (Suppl. Fig. 1. E): this optimal matching sequence is called the warping path (Suppl. Fig. 1. B). When comparing identical time-series the warping path exactly follows the main diagonal (no elongation steps are necessary), while phase-differences between signals will introduce elongation steps, therefore the length of the warping path will increase compared to the main diagonal. The difference between the warping path length and the length of the main diagonal can characterize the overall phase difference and the stability of this time-delay structure between the two time-series (Meszlényi et al., 2016a) and this measure is referred as warping path length throughout the paper.

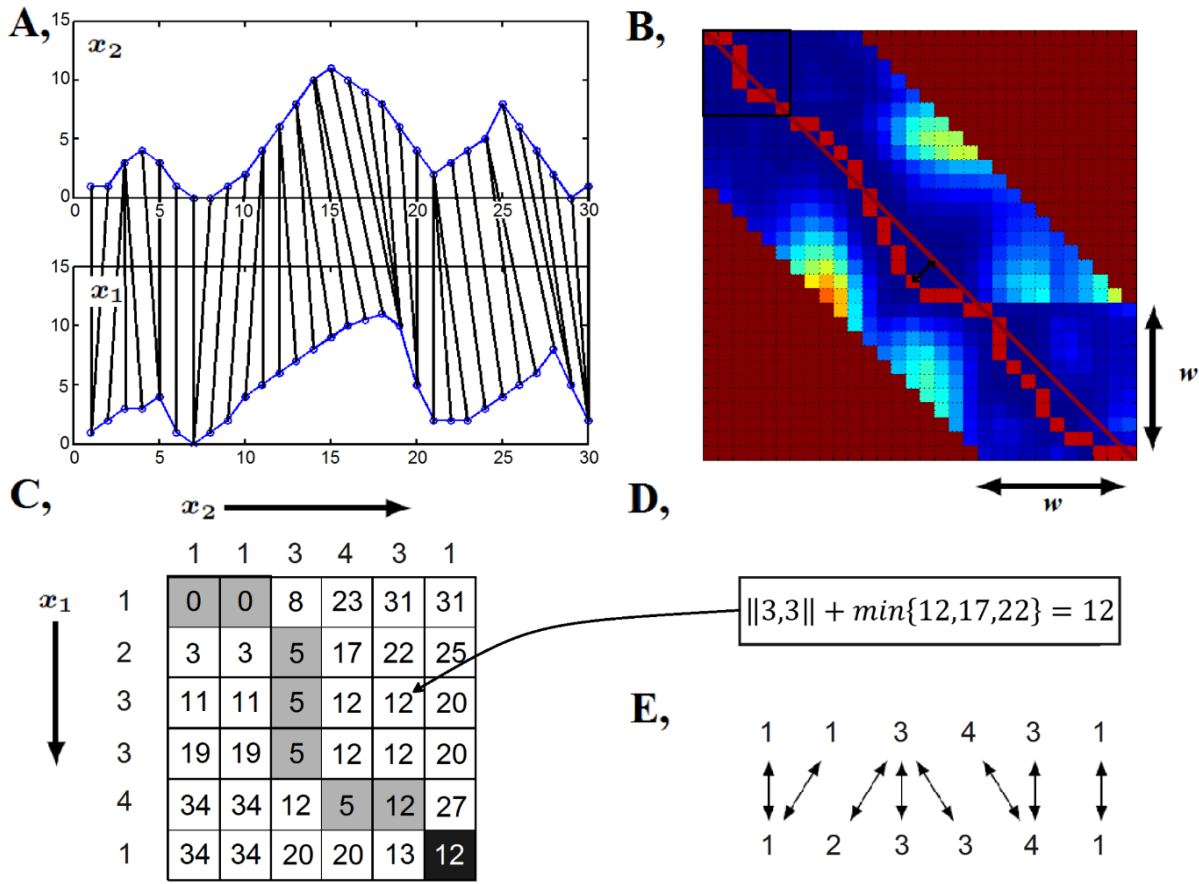

*Supplementary Figure 1.: Figure adapted from (Meszlényi et al., 2017) A, $x_1$ and $x_2$ time series compared with DTW: the i-th element of $x_1$ is elastically matched to the appropriate element of $x_2$. B, The filled-out DTW matrix plotted as a heat-map (hot colors represent larger values), w denotes the size of the warping window, the maximal allowed time-lag between two matched time series element. The main diagonal is represented by the dark red line, while the optimal warping path is plotted with red. The time-delay between the $x_1$ and $x_2$ time series at a given time-point is given by the warping path's deviation from the main diagonal (represented by the black arrow) C, Calculation of DTW distance by filling out the DTW matrix: the example shows the first six element of $x_1$ and $x_2$ time series (highlighted with the black rectangle in Fig… .B). Elements of $x_1$ corresponds to rows, while elements of $x_2$ corresponds to columns of the matrix. The optimal warping path is highlighted with dark grey. D, Formula to calculate entry (i,j) – in this example entry (3,5): squared distance of $x_1(i)$ and $x_2(j)$ plus the minimum of the matrix entries (i-1,j), (i-1,j-1), (i,j-1) E, Optimal matching of the first six elements of $x_1$ and $x_2$ revealed by the DTW matrix.*

## 9.1 Results of the traditional machine learning methods

We also tested how traditional machine learning methods perform compared to the CCNN method. We conducted experiments with two algorithms that can handle the curse of dimensionality with feature selection, namely a linear SVM classifier combined with ANOVA F-test based feature selection described in (Abraham et al., 2014) and a LASSO model also frequently used in fMRI based classification (Li* et al., 2009; Rosa et al., 2015; Ryali et al., 2010). Our adopted approach is detailed in (Meszlényi et al., 2016b).

## 9.2 Results on the combined DTW distance and warping path length dataset

As the CCNN method's best result was achieved on the combination of two connectivity features, DTW distance and path length, we calculated baseline accuracies on this dataset with different hyper-parameter setups of both SVM and LASSO models. Based on the results presented in Supplementary Table 1 and 2 we can conclude that neither of these methods could achieve better performance than the CCNN model. We also note that the resulting predictions from different hyper-parameter setups do not differ significantly ($p>0.15$ in all cases).

### 9.2.1 SVM results

As all our neural network models (including the CCNN architecture) are trained to extract 128 features from the connectivity data, we selected the best 128 connectivity features based on the ANOVA F-tests, and performed linear SVM classification on the selected features with five different values of the complexity parameter C. The accuracy values are summarized in Supplementary Table 1:

| C | 1 | 0.5 | 0.1 | 0.05 | 0.01 |
|---|---|---|---|---|---|
| Accuracy (%) | 62.3 | 61.6 | 66.4 | 66.4 | 63.0 |

*Supplementary Table 1: Results of the linear SVM classification*

### 9.2.2 LASSO results

In case of the LASSO model based classification, we ran experiments with six values of the regularization hyper-parameter $\lambda$ to enforce the selection of approximately 128 connectivity features. Supplementary Table 2 summarizes the accuracy values of the resulting classifications as well as the mean and standard deviation of the number of selected features in the seven folds of cross-validation.

| $\lambda$ | 0.001 | 0.0009 | 0.0008 | 0.0007 | 0.0006 | 0.0005 |
|---|---|---|---|---|---|---|
| No. features (mean±std) | 119±2.1 | 121±2.6 | 122±3.5 | 123±3.5 | 125±3.4 | 127±4.4 |
| Accuracy (%) | 69.2 | 69.9 | 69.9 | 69.2 | 68.5 | 69.2 |

*Supplementary Table 2: Results of the LASSO classification*

## 9.3 Results on the single datasets

The CCNN method achieved considerably better accuracy on the combined dataset than on any dataset containing only a single connectivity feature, therefore it is also important to examine whether SVM and LASSO models can similarly well utilize information from multiple connectivity features. Consequently we conducted classification experiments with SVM and LASSO models based on the three single connectivity feature datasets: correlation, DTW distance and warping path length.

### 9.3.1 SVM Results

Similarly to the training combined dataset, we selected the best 128 connectivity features based on the ANOVA F-tests, and performed linear SVM classification on the selected features with C = 0.05. The accuracy values are summarized in Supplementary Table 3:

| C=0.05 | CORR | DTW | Path length | *DTW+Path length* |
|---|---|---|---|---|
| Accuracy (%) | 54.1 | 67.1 | 64.4 | *66.4* |

*Supplementary Table 3: Results of the linear SVM classification*

### 9.3.2 LASSO results

We performed LASSO classification on the three single dataset with $\lambda$ = 0.0008. The results are summarized in Supplementary Table 4:

| $\lambda$ =0.0008 | CORR | DTW | Path length | *DTW+Path length* |
|---|---|---|---|---|
| Accuracy (%) | 60.3 | 59.6 | 69.9 | *69.9* |

*Supplementary Table 4: Results of the LASSO classification*

## 9.4 Discussion of the results of the traditional algorithms

On the combined DTW distance and warping path length dataset the best SVM accuracy is 5.5% lower than the CCNN result, while in case of the LASSO classifiers, the best performing model has 2% lower accuracy than the CCNN model, however neither of these differences are significant (p>0.1). Since we ran experiments with different hyper-parameters, our estimates of these test-performances might be over-optimistic. Even with this slight positive bias none of these classifiers achieved better results than the CCNN model.

Most importantly we note that neither the SVM, nor the LASSO model was able to utilize additional information from the combination of different connectivity features, i.e. the SVM classifier even achieved slightly better (0.6% higher) accuracy on the single DTW distance dataset than on the combined data and the LASSO model has the same result on the single path length and the combined datasets.

Our results confirm that while the convolutional model is able to integrate information from multiple different connectivity features, traditional machine learning algorithms do not gain performance from the additional information. Furthermore, based on current deep learning research one can assume that both with growing number of measurements and with growing number of measured dimensions, i.e. in case of increasingly complex data, the difference between the performance of traditional machine learning models and deep learning methods increase (Goodfellow et al., 2016; LeCun et al., 2015). Therefore the proposed CCNN architecture can hold great potential for future research.

## 9.5 Supplementary Figure

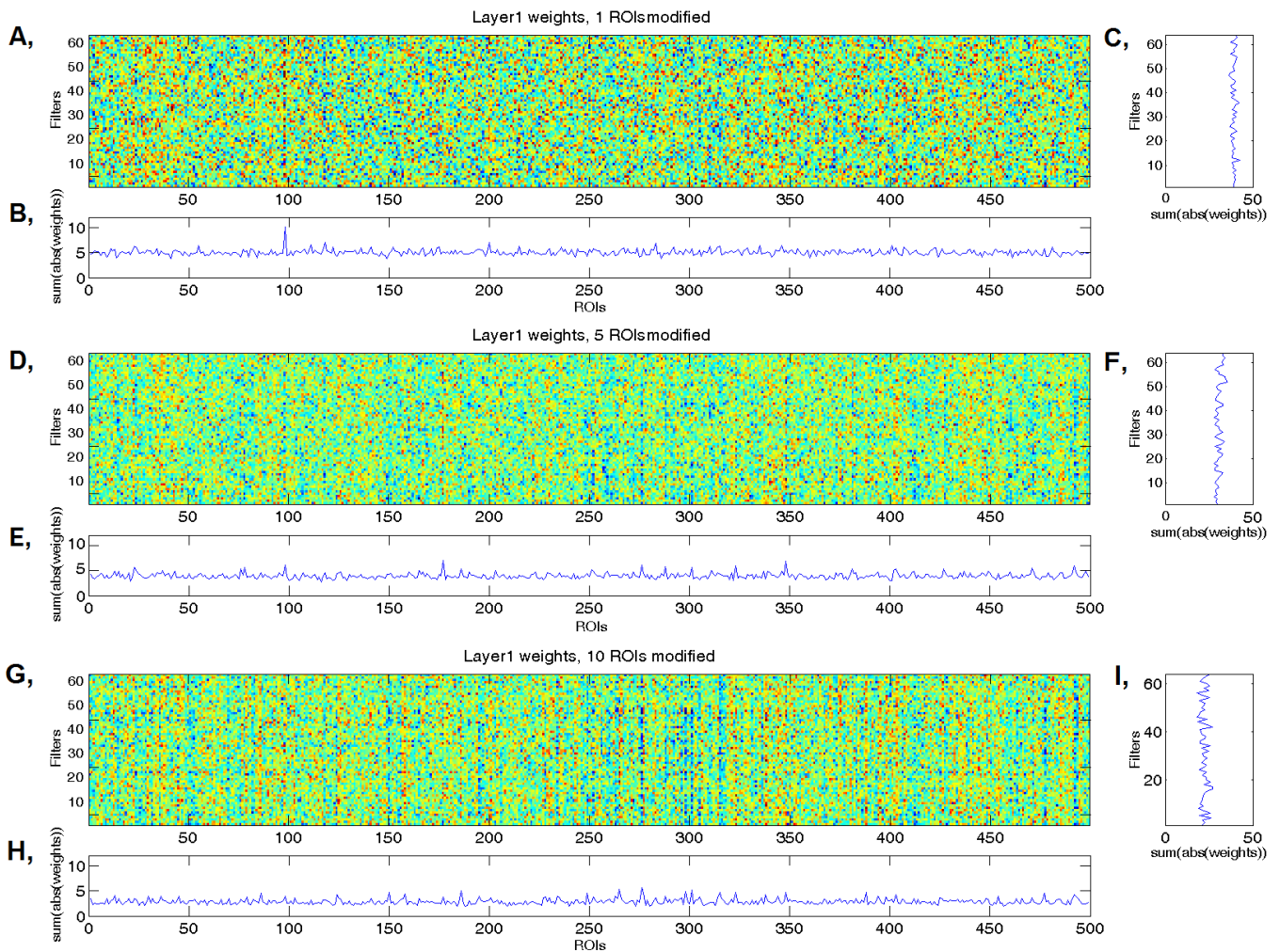

*Supplementary Figure 2: A-C, Learned weights of the first convolutional layer of the CCNN model trained on the whole simulated dataset, with one ROI modification. A, Colormap of the 64x499 weights of the first convolutional layer. B, To determine which ROIs play important role in the classification, we summarized the absolute values of the weights through the 64 filters. High values represent ROIs that have significant effect in most filters: the sole highest peak is indeed identical to the one altered ROI C, To determine which filters are the most effective, we summarized the absolute values of weights through the 499 ROIs. High values represent filters that have substantial influence on the output. D-E, Learned weights of the first convolutional layer of the CCNN model trained on the whole simulated dataset, with five ROI modification. D, Colormap of the 64x499 weights of the first convolutional layer. E, Summarized absolute values of the weights through the 64 filters. Four of the five highest peaks overlap with the modified five ROIs. F, Summarized absolute values of weights through the 499 ROIs. G-I, Learned weights of the first convolutional layer of the CCNN model trained on the whole simulated dataset, with ten ROI modification. G, Colormap of the 64x499 weights of the first convolutional layer. H, Summarized absolute values of the weights through the 64 filters. Four of the ten highest peaks overlap with the modified ROIs. I, Summarized absolute values of weights through the 499 ROIs.*

## 9.6 Supplementary Material - References